 \pdfoutput=1 
\documentclass{article}
\usepackage{ijcai24}

\usepackage{times}
\usepackage{tikz}
\usetikzlibrary{shapes,arrows,positioning}
\usepackage{soul}
\usepackage{url}
\usepackage[hidelinks]{hyperref}
\usepackage[utf8]{inputenc}
\usepackage[small]{caption}
\usepackage{graphicx}
\usepackage{amsmath}
\usepackage{amsthm}
\usepackage{booktabs}
\usepackage{algorithm}
\usepackage{algorithmic}
\usepackage[switch]{lineno}

\newtheorem{definition}{Definition} 
\usepackage{amsfonts} 
\usepackage{amssymb}
\usepackage{pifont}
\DeclareMathOperator*{\argmax}{arg\,max}

\title{Building Expressive and Tractable Probabilistic Generative Models: A Review}

\author{
Sahil Sidheekh
\and
Sriraam Natarajan
\affiliations
The University of Texas at Dallas\\
\emails
\{sahil.sidheekh, sriraam.natarajan\}@utdallas.edu,
}

\begin{document}

\maketitle

\begin{abstract}
    We present a comprehensive survey of the advancements  and  techniques in the field of tractable probabilistic generative modeling, primarily focusing on Probabilistic Circuits (PCs). We provide a unified perspective on the inherent trade-offs between expressivity and tractability, highlighting the design principles and algorithmic extensions that have enabled building expressive and efficient PCs, and provide a taxonomy of the field. We also discuss recent efforts to build deep and hybrid PCs by fusing notions from deep neural models, and outline the challenges and open questions that can guide future research in this evolving field. 
\end{abstract}

\section{Introduction}

Generative modeling plays an important role in the fields of machine learning and artificial intelligence, as it offers a powerful toolkit for understanding, interpreting as well as recreating complex patterns present in our data-rich world. By employing probability theory as a principled way to capture the inherent uncertainty in a given dataset, these models aim to approximate the underlying distribution or random process responsible for generating the data. Consequently, probabilistic generative models possess the potential to solve a diverse range of problems, including generating new data samples, performing inference given observations, estimating likelihood of events, and reasoning about uncertain information.

However, learning the distribution back from data is a challenging problem that often necessitates trade-offs between modeling flexibility and the tractability of probabilistic inference. Early generative models prioritized enabling tractable inference, often by imposing probabilistic structures over random variables, in the form of graphical models \cite{koller2009probabilistic}. However, as a result, they lacked the flexibility to model complex distributions. The field of {\em Tractable Probabilistic Models} (TPMs) have since evolved,
with expressive parameterizations and learning paradigms proposed, resulting in a broad and popular class of models under the unified notion of \textit{probabilistic circuits}. Designed from the perspective of tractability, these models enable efficient inference and exact probabilistic reasoning, making them suitable for tasks demanding fast and exact computations. However, they still struggle to capture dependencies when the data complexity and dimensionality increase. 

In contrast, advancements in deep learning have given rise to expressive Deep Generative Models (DGMs) that exploit the power of neural networks to learn flexible representations of complex data distributions. Notable examples include generative adversarial networks, variational autoencoders, normalizing flows and diffusion models. Prioritizing expressiveness, these models have demonstrated impressive proficiency in capturing complex dependencies and generating high fidelity samples. However, unlike TPMs, they often lack the ability to reason explicitly about the learned distribution.

Bridging the gap between TPMs and DGMs is thus a fascinating area of research, aiming to combine their strengths and create hybrid models that are expressive as well as  tractable. This survey aims to comprehensively explore techniques and recent advancements in this direction. While previous surveys have extensively studied DGMs \cite{taylor2021deep} and TPMs \cite{sanchez2021sum} independently, analyzing their design principles and associated challenges, a unified and cohesive view is still lacking. Through this work, we thus aim to fill this gap and provide researchers with a holistic understanding of the field. We hope to highlight the benefits and challenges of this synergistic combination to motivate enhanced research in this direction.
 
We will begin by discussing the building blocks, properties, learning methodologies as well as challenges when building tractable generative models, focusing on probabilistic circuits, and provide a broad taxonomy of the field. 
We will then discuss hybrid techniques that merge TPMs with DGMs to achieve the best of both worlds. Finally, we will identify challenges, open problems, and potential directions that can lay the foundations for future research in this field. We aim to serve the dual purpose of providing beginners interested in the field with a broad understanding of how to build expressive and tractable generative models, while empowering experienced researchers to push the boundaries by understanding the intricacies and complexities of the domain.

\paragraph{Notation.} We use $\mathbf{X}=(X_1, X_2, \ldots, X_d)$ to denote a set of $d-$ random variables and $\mathbf{x} = ( x_1, x_2, \ldots, x_d)$ denote an assignment of values to $\mathbf{X}$. We use $P_\mathbf{X}$ to denote the true probability distribution over $\mathbf{X}$ and $P_{\theta}$ to denote the distribution captured by a generative model parameterized by $\theta$.


\section{Tractable Probabilistic Modeling}

Given a dataset $\mathcal{D} = \{\mathbf{x}^{(i)} \}_{i=1}^{M}$ of $M$ instances, each comprising $d$ features, the generative modeling problem can be formalized as learning a model $\theta$ that best explains $\mathcal{D}$, i.e. finding an optimal $\theta^*$ such that $P_{\theta^*} \approx P_{\mathbf{X}}$. The optimization problem can be formulated using maximum likelihood as: 
$
    \theta^* = \underset{\theta}{\argmax} \ P_{\theta}(\mathcal{D}) = \underset{\theta}{\argmax} \prod_{i=1}^{M} P_{\theta}(\mathbf{x}^{(i)}).
$
A probabilistic generative model is said to be \textit{tractable} for an inference task, if the answer to the corresponding probabilistic query can be computed \emph{exactly} in time that is polynomial with respect to the size of the model and the input. Thus, tractability is a spectrum that depends not only on the characteristics of the generative model, but also on the type of probabilistic query, i.e. the inference task. As we will see, some inference tasks are inherently harder than others and a model tractable for one query might not be for another.

\subsection{Inference Queries} 
We begin by outlining some of the prevalent inference tasks that we might be interested in performing, in order to make decisions using a probabilistic generative model effectively.
The most frequent and basic inference scenario typically involves calculating the probability associated with a specific assignment of values to all the random variables. This is commonly referred to as \textbf{evidential inference} and involves computing $P_{\theta}(\mathbf{X}=\mathbf{x})$ for a given assignment $\mathbf{x}$ exactly, without resorting to any approximations.
In many real world problems, we are interested in reasoning about only a subset of the modeled variables due to the non-homogenous nature of the data or missing features. Or sometimes, in the presence of sensitive features, enforcing fairness in decision-making might require us to marginalize out the effect of certain variables. More formally, given subsets $\mathbf{X}_1,\mathbf{X}_2$ such that $\mathbf{X}_1 \cup \mathbf{X}_2 = \mathbf{X}$ and $\mathbf{X}_1 \cap \mathbf{X}_2 = \emptyset$, evaluating the likelihood of the subset of variables of interest (say $\mathbf{X}_1$) by marginalizing out rest is referred to as \textbf{marginal inference}, i.e. it involves computing
$
    P_{\theta}(\mathbf{X}_1) = \int_{\mathbf{X}_2}P_{\theta}(\mathbf{X_1},\mathbf{X}_2)d\mathbf{X_2}.
$
Marginal inference involves performing evidential inference as a subroutine a large (possibly infinite) no. of times, and is thus a harder task than the latter.  
The third type of inference scenario involves computing the probability of an event $A$, given that another event $B$ has happened. This is referred to as \textbf{conditional inference}, and it involves computing 
$
    P_{\theta}(\mathbf{X}_1|\mathbf{X}_2) = \frac{P_{\theta}(\mathbf{X}_1,\mathbf{X}_2) }{P_{\theta}(\mathbf{X}_2) } = \frac{P_{\theta}(\mathbf{X}_1,\mathbf{X}_2) }{ \int_{\mathbf{X}_1}P_{\theta}(\mathbf{X_1},\mathbf{X}_2)d\mathbf{X_1}}.
$\\
\textbf{Maximum-a-Posteriori (MAP) inference} seeks the most probable variable assignments based on evidence, essentially maximizing the posterior distribution over the variables of interest. This task is crucial for parameter estimation and predicting the most likely outcomes. More formally, it invoves computing
 $\argmax_{\mathbf{X}_1} P_{\theta}(\mathbf{X}_1|\mathbf{X}_2) = \argmax_{\mathbf{X}_1} P_{\theta}(\mathbf{X}_1,\mathbf{X}_2)$.  
\\

\subsection{Enforcing Tractability by Imposing Structure}
A prevalent strategy to attain tractability in probabilistic inference for generative models is the imposition of structural constraints within them.  For instance, considering all random variables as independent leads to a highly structured model: $P_{\theta}(\mathbf{X})=\prod_{i=1}^{d}P_{\theta}(X_i)$. This simplifies the computation of marginals, conditionals, and MAP, reducing them to operations over univariate distributions. Incorporating such factorizations of the joint distribution is thus key for achieving tractability. However, this often limits the model's expressiveness, as more restrictive assumptions reduce the variety of representable probability distributions. To balance flexibility and simplicity, \emph{mixture models}, which are convex combinations of simpler distributions, are typically employed. A notable example is the Gaussian Mixture Model, which is theoretically capable of approximating any continuous distribution in the limit, given adequate components
\cite{lindsay1995mixture}.
Thus, designing algorithms that enforce structure via factorizations and flexibility via mixtures is an effective approach to build tractable models without sacrificing expressivity.

\definecolor{header}{RGB}{220, 235, 250}
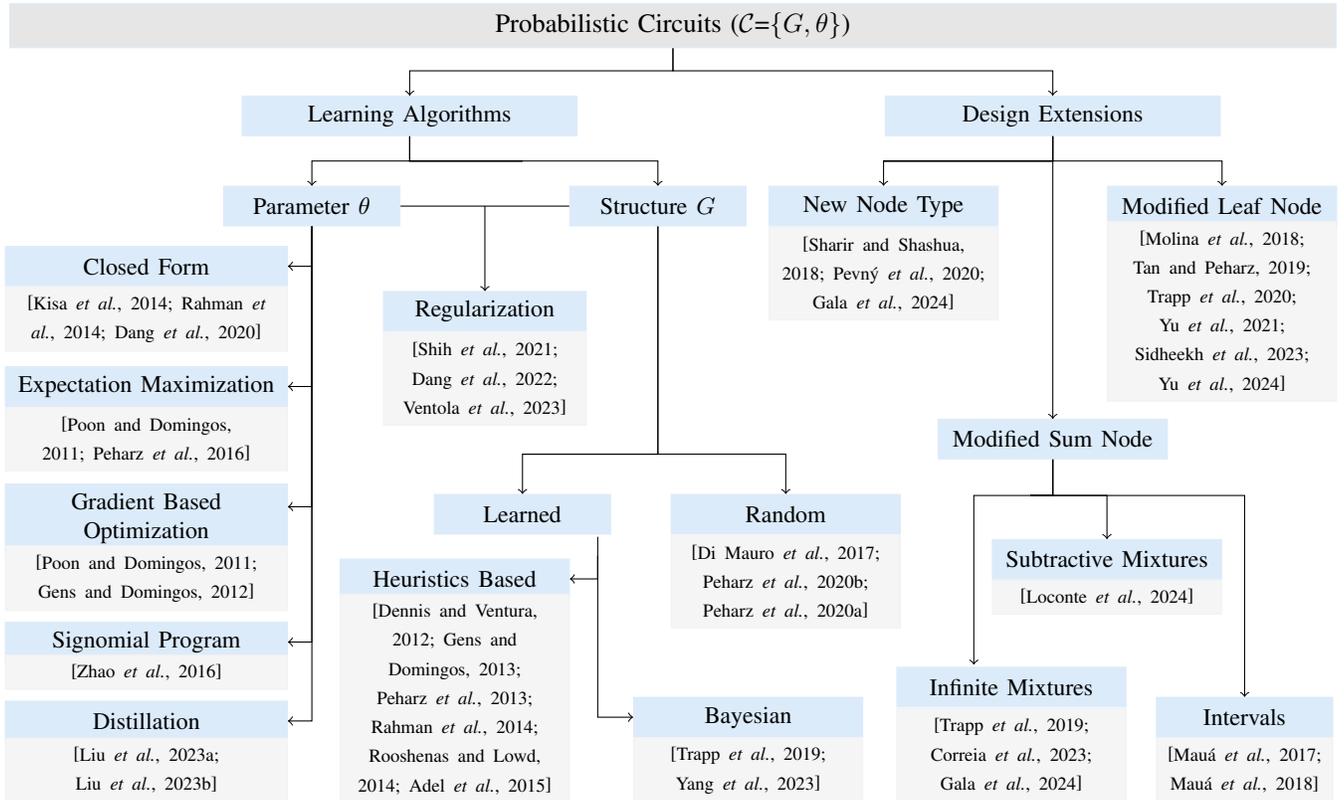
\begin{figure*}[t]
    \centering
\begin{tikzpicture}[
    block/.style={
    rectangle,
    draw=header,
    fill=header!100,
    text centered,
    text width=3cm,
    minimum height=1.5em},
    leaf/.style={
    rectangle,
    draw=gray!10,
    fill=gray!8,
    text centered,
    text width=3cm,
    minimum height=1em},
    line/.style={draw, -latex'}
]
\node [block, fill=white!5, text width=0em,  minimum height=0em, draw=white] (anchor1)  at (0,0) {};
\node [block, fill=gray!20, text width=49.5em] (root)  at (7,0) { Probabilistic Circuits ($\mathcal{C}$=$\{ G, \theta\}$)};

\node [block, text width=12em] (algo) at (3.5,-1.2) {\small Learning Algorithms};
\node [block, text width=12em] (design) at (12.05,-1.2) {\small Design Extensions};

\node [block, text width=6em] (param) at (2.2,-2.4) {\small Parameter $\theta$};
\node [block, text width=10em] (closed) at (0,-3.2) {\small Closed Form};
\node [leaf, align=center, text width=10em] (cite-closed) at (0,-3.9) {\scriptsize 
\cite{kisa2014probabilistic,rahman2014cutset,dang2020strudel}};

\node [block, text width=7em] (regularization) at (4.5,-3.8) {\small Regularization};
\node [leaf, align=center, text width=7em] (cite-regularization) at (4.5,-4.7) {\scriptsize 
\cite{shih2021hyperspn,dang2022sparse,ventola2023probabilistic}};

\node [block, text width=10em] (em) at (0,-4.8) {\small Expectation Maximization};
\node [leaf, align=center, text width=10em] (cite-em) at (0,-5.5) {\scriptsize 
\cite{poon2011spn,peharz2016latentvariablespn}};

\node [block, text width=10em] (gradient) at (0,-6.55) {\small Gradient Based Optimization};
\node [leaf, align=center, text width=10em] (cite-gradient) at (0,-7.35) {\scriptsize 
\cite{poon2011spn,gens2012discriminative}};

\node [block, text width=10em] (signomial) at (0,-8.2) {\small Signomial Program};
\node [leaf, align=center, text width=10em] (cite-signomial) at (0,-8.6) {\scriptsize 
\cite{zhao2016unified}};

\node [block, text width=10em] (distillation) at (0,-9.25) {\small Distillation};
\node [leaf, align=center, text width=10em] (cite-distillation) at (0,-9.9) {\scriptsize 
\cite{liu2023scaling,liu2023understanding}};

\node [block, text width=6em] (structure) at (6.8,-2.4) {\small Structure $G$};
\node [block, text width=6em] (learned) at (5,-6.5) {\small Learned};
\node [block, text width=8em] (random) at (8.5,-6.5) {\small Random};
\node [leaf, align=center, text width=8em] (cite-random) at (8.5,-7.4) {\scriptsize 
\cite{di2017fast,peharz20a-rat-spn,peharz_20_einsum}};

\node [block, text width=8em] (heuristic) at (4.1,-7.36) {\small Heuristics Based};
\node [leaf, text width=8em] (cite-heuristic) at (4.1,-8.95) {\scriptsize \cite{ventura12buildspn,gens2013learnspn,peharz2013greedy,rahman2014cutset,rooshenas14learning,adel2015learning} };

\node [block, text width=8em] (bayesian) at (8,-9.2) {\small Bayesian};
\node [leaf, text width=8em] (cite-bayesian) at (8,-9.9) {\scriptsize \cite{trapp2019bayesian,yang23bayesian}};

\node [block, text width=8em] (newnode) at (9.8,-2.4) {\small New Node Type};
\node [leaf, text width=8em] (cite-newnode) at (9.8,-3.3) {\scriptsize \cite{sharir2018sum,sptn,gala2024probabilistic}};

\node [block, text width=8em] (modifiedsum) at (12.05,-5.5) {\small Modified Sum Node};
\node [block, text width=8em] (infinite) at (11.5,-8.8) {\small Infinite Mixtures};
\node [leaf, text width=8em] (cite-infinite) at (11.5,-9.7) {\scriptsize \cite{trapp2019bayesian,correia2023continuous,gala2024probabilistic}};

\node [block, text width=8em] (negativemixture) at (12.77,-7.1) {\small Subtractive Mixtures};
\node [leaf, text width=8em] (cite-negativemixture) at (12.77,-7.6) {\scriptsize \cite{loconte2024subtractive}};

\node [block, text width=6em] (interval) at (14.6,-9.2) {\small Intervals};
\node [leaf, text width=6em] (cite-interval) at (14.6,-9.9) {\scriptsize \cite{maua2017credal,maua2018robustifying}};

\node [block, text width=8em] (modifiedleaf) at (14.3,-2.4) {\small Modified Leaf Node};
\node [leaf, text width=8em] (cite-modifiedleaf) at (14.3,-3.8) {\scriptsize \cite{molina2018mixed,tan2019hierarchical,trapp2020deepgaussianprocess,yu2021uai_momogps,sidheekh2023probabilistic,yu2024characteristic}};

\draw[->] (root)  |- (7, -0.6) -|   (algo);
\draw[->] (root)  |- (7, -0.6) -|   (design);
\draw[->] (param) |- (1.9, -3.2) --   (closed);
\draw[->] (param) |- (1.9, -4.8) --   (em);
\draw[->] (param) |- (1.9, -6.4) --   (gradient);
\draw[->] (param) |- (1.9, -8.2) --   (signomial);
\draw[->] (param) |- (1.9, -9.25) --   (distillation);
\draw[] (param) -- (structure);
\draw[->] (4.5,-2.4) -- (regularization);
\draw[->] (structure) |- (5, -5.7) -|   (learned);
\draw[->] (structure) |- (5, -5.7) -|   (random);
\draw[->] (6, -7.05)    |-   (heuristic);
\draw[->] (6, -6.8)    |-   (bayesian);

\draw[->] (design)    |- (9.8,-1.8) -| (newnode);
\draw[->] (design)    |- (9.8,-1.8) -| (modifiedleaf);
\draw[->] (design)    |- (12.05,-1.8) --   (modifiedsum);

\draw[->] (algo)    |- (5,-1.8) -| (param);
\draw[->] (algo)    |- (5,-1.8) -| (structure);

\draw[->] (modifiedsum)    |- (11.,-6.25) --  (11.,-8.5);
\draw[->] (modifiedsum)    |- (11.75,-6.25) -|   (interval);
\draw[->] (modifiedsum)    |- (12.25,-6.25) -| (negativemixture);

\end{tikzpicture}
\caption{A broad taxonomy of the literature on improving probabilistic circuits via better learning algorithms and design extensions.}
\label{pc-taxonomy}
\end{figure*}

\section{Probabilistic Circuits}
Building on the above principle of factorizations and mixtures, several classes of tractable probabilistic models have emerged, such as Arithmetic Circuits \cite{darwiche2003differential-ac}, Probabilistic Sentential Decision Diagrams \cite{kisa2014probabilistic}, AND-OR search spaces \cite{marinescu2005and}, Sum Product Networks \cite{poon2011spn}, Cutset Networks \cite{rahman2014cutset}, etc. Recently, \cite{ProbCirc20} presented a unified view of such models, generalizing them under the umbrella notion of \textit{Probabilistic Circuits} (PCs). We emphasize on PCs as a tractable representation for learning data distributions {\bf but go beyond their work by providing a unified view of the different models, their learning methodologies, and presenting the latest work in employing ideas from deep learning}.


\begin{definition}
    A Probabilistic Circuit $\mathcal{C}$ is a computational graph that compactly encodes a probability distribution via factorizations and mixtures. It consists of three types of nodes - Sums, Products and Leaf Distributions. Each node in the graph computes a non-negative function, which can be interpreted as an un-normalized probability measure over a subset of random variables, which is referred to as the scope of the node. The computational graph is evaluated bottom up and is recursively defined as follows:
    \begin{itemize}
        \item For a sum node $(+)$ with scope $\psi_{+} \subseteq \mathbf{X}$, the output is defined as a convex combination of the outputs of its children. i.e $+(\mathbf{x}_{\psi_{+}}) = \sum_{N_i \in \text{child}(+)} w_iN_i(\mathbf{x})$, where $w_i \geq 0$ and $\sum_i w_i = 1$.

        \item For a product node $(\times)$ with scope $\psi_{\times} \subseteq \mathbf{X}$, the output is defined as a product of the outputs of its children. i.e $\times(\mathbf{x}_{\psi_{\times}}) = \prod_{N_i \in \text{child}(\times)} N_i(\mathbf{x}_{\psi_{N_i}})$.

        \item A leaf node $L$ represents a simple tractable univariate distribution over its scope $\psi_{L} \subseteq \mathbf{X}$ , such as a Gaussian, and its output is defined as the probability density (or mass for discrete variables) under the corresponding distribution. $L(\mathbf{x}_{\psi_L})=P_{L}(\mathbf{x}_{\psi_L})$
    \end{itemize}
    The output of the root node constitutes the modeled density.
\end{definition}
A probabilistic circuit $\mathcal{C} = \{ \mathcal{G}, \theta \} $ thus has both a structure encoded as a computational graph $\mathcal{G}$ and parameters $\theta$ which corresponds to the weights associated with the sum nodes and the parameters of the leaf distribution. It can be viewed as a hierarchical mixture model that compactly encodes an exponential no. of factorized mixture components.

\subsection{Structural Properties for Probabilistic Circuits}
 In order to ensure that $\mathcal{C}$ models a valid distribution and supports tractability, we need to define further structural properties over them, which we elaborate below.
\begin{definition}[Smoothness]
    A probabilistic circuit is said to be \textbf{smooth}, if all of it's sum nodes are defined over children having the same scope.
\end{definition}
\begin{definition}[Decomposability]
    A probabilistic circuit is said to be \textbf{decomposable}, if all of it's product nodes are defined over children having disjoint scope. 
\end{definition}
\begin{definition}[Determinism]
    A probabilistic circuit is said to be \textbf{deterministic}, if, for all of its sum nodes, the output of at most one of its children is non-zero for any given input.
\end{definition}

Intuitively, smoothness implies that the sum nodes represent valid mixture distributions. This, in turn implies the tractability of evidential inference, as the data density can be computed by evaluating the circuit bottom up, which can be done in time linear in the size of the circuit.
Further, circuits that are smooth and decomposable additionally support tractable computation of marginal and conditional queries. This is because the integrals involved in these queries decomposes over the sum and product nodes to their children.
Thus for a smooth and decomposable PC, we can push down the integrals recursively until we reach the leaves. Since the leaf distributions are simple and tractable, the integrals can be computed analytically. Evaluating the marginal or conditional query then reduces to performing a bottom-up pass on the circuit, with the values of the leaf nodes set to their corresponding integrals, and is thus also linear in the size of the circuit. However, smoothness and decomposability are not sufficient for tractably computing MAP queries. This is because the maximizer over a convex combination of distributions is not necessarily the convex combination of the maximizers of the individual distributions. 
However, we can perform MAP inference tractably over smooth, decomposable, and \emph{deterministic} PCs. Note that deterministic PCs are also sometimes called \textit{selective} as the sum nodes can be viewed as selecting one of its children
\cite{peharz2014learning}.


\subsection{Parameter Learning}
As PCs support density evaluation, we can employ maximum likelihood to learn their parameters. In the presence of determinism, the output of the root node reduces to a weighted product of simple factorized distributions, and the parameters can be estimated in closed form \cite{rahman2014cutset}. However, in the more general case of smooth and decomposable circuits, a closed form solution is not available, and we will have to resort to iterative optimization schemes to maximize the data likelihood. 

\paragraph{Gradient Based Optimization.} Since PCs are defined as computational graphs, they are differentiable, and the partial derivative of the output of the root node (which represents the data density) w.r.t to each node in the circuit can be easily computed. 
Thus, initializing the model parameters randomly, these gradients can then be used to iteratively update the parameters to maximize the likelihood and learn the data distribution. 
In practice, one could implement these computational graphs using packages that support automatic differentiation, and backpropagate the gradients efficiently.
While one can employ full batch gradient ascent to optimize this objective, its stochastic version that samples mini batches is more compute efficient and fast, and is popularly used for learning PCs in the generative and well as discriminative settings \cite{poon2011spn,gens2012discriminative}. 
More complex gradient-based optimizers such as Adam 
that incorporates momentum have also been used to learn PCs \cite{peharz20a-rat-spn,peharz_20_einsum,sidheekh2023probabilistic}.

\paragraph{Expectation Maximization.}
An alternative optimization scheme proposed for PCs is expectation-maximization (EM), which is popularly used for maximum likelihood learning in the presence of missing data \cite{dempster1977maximum}. The crux of EM involves iterating between the following two steps until convergence, after initializing from a random configuration - (1) The E-Step, which involves computing the expected value of the missing variables given the observed variables and (2) the M-step, which maximizes the likelihood given the complete expected assignment. As PCs are essentially hierarchical mixture models, their sum nodes can be viewed as marginalizing out an unobserved discrete latent variable.
\cite{peharz2016latentvariablespn} formalized this latent variable interpretation for PCs by explicitly introducing the unobserved latent variables in the computational graph to create an \textit{augmented} circuit. As we do not know the assignments for these variables for the data points in hand, it becomes a learning with missing data problem and can be solved using expectation maximization. Compared to gradient ascent, EM has been observed to result in a larger boost in the data likelihood \cite{peharz2016latentvariablespn,peharz20a-rat-spn,peharz_20_einsum} especially in the early phase of learning.

\cite{zhao2016unified} presented a unified view of parameter learning in smooth and decomposable PCs. They observed that the MLE optimization problem can be formulated as a signomial program and proposed two algorithms for parameter learning - (1) sequential
monomial approximations (SMA) that generalized gradient descent and (2) the concave-convex procedure (CCP) that generalized expectation maximization. All of the approaches discussed above focus on data-driven parameter learning, which can be challenging when the data is \emph{noisy} and \emph{sparse}. Knowledge-intensive learning of PCs \cite{mathur2023knowledge,karanam2024unified}  has been proposed as a more effective and robust framework under such adverse settings.

\subsection{Structure Learning}
 The computational graph structure $\mathcal{G}$ of a PC encodes the factorizations of the joint distribution, and hence can impact its expressivity. However, the optimal structure can vary for each data distribution and is seldom known apriori.

\paragraph{Heuristics Based.}
Learning the graph structure from data was first explored by \cite{ventura12buildspn}. They observed that the latent variable associated with a sum node should help explain interactions between variables in its scope. Hence they proposed to strategically locate sum nodes over groups of variables that have pronounced interdependencies via clustering. Subsequently, \cite{gens2013learnspn} showed that the above apporach does not make use of context specific independencies and is prone to splitting highly dependent variables into different clusters, thus causing a large loss of likelihood. Instead they proposed a greedy iterative algorithm, known as LearnSPN that could be used to define the scopes for both the sum nodes as well as product nodes. Specifically, at a product node, they employed statistical independence testing to identify mutually independent subsets of variables. And at a sum node, they employed EM based clustering to group similar instances as the support for each of its child. The weight for a sum node edge could now be defined as the fraction of data points belonging to its corresponding cluster. Recursively employed, this algorithm resulted in the extraction of sets of columns from the data matrix at a product node and sets of rows at a sum node until a univariate leaf was reached. Building further, \cite{rooshenas14learning} proposed to merge the indirect interactions modeled by the latent variables clustered at higher levels, with direct interactions of observed variables by employing tractable Markov networks at the lower levels.  Similarly, \cite{adel2015learning} proposed an SVD based structure learning algorithm that merged the row-wise and column-wise splitting of the data matrix employed by LearnSPN into a single operation of extracting rank-one submatrices. Other approaches have explored incorporating information bottleneck \cite{peharz2013greedy} as well as cutset-conditioning that mimicks decision tree learning \cite{rahman2014cutset}. However all of the above methods are based on heuristics and lacks solid theoretical grounding.

\paragraph{Bayesian Approaches.}  Perhaps the most principled and elegant approach to structure learning for PCs is to adopt a Bayesian view. By viewing the parameterized density as a function of both $\theta$ and $\mathcal{G}$, i.e. $P_{\theta, \mathcal{G}}(\mathbf{x}) = P(\mathbf{x}|\theta, \mathcal{G})$, we can define \textit{Bayesian Structure Score} ($\mathcal{B}$) as the contribution of $\mathcal{G}$ to the overall likelihood, i.e.:
$
    \mathcal{B}(\mathcal{G}) = P(\mathcal{D}|\mathcal{G}) = \int_{\theta} P(\mathcal{D}|\theta, \mathcal{G})P(\theta|\mathcal{G})d\theta
 = \int_{\theta} P(\theta|\mathcal{G}) \prod_{\mathbf{x} \in \mathcal{D}} P(\mathbf{x}|\theta, \mathcal{G}) d\theta
$
\\
Intuitively, this reduces to assuming a prior distribution $P(\theta|\mathcal{G})$ over the parameters and computing the contribution of the structure $\mathcal{G}$ to the likelihood by marginalizing out the parameters. Such a score can then be used to optimize over structures in a Bayesian way \cite{friedman2003being} by employing search algorithms
\cite{russell2010artificial} 
or even structural expectation maximization \cite{Friedman1998TheBS}. However, computing the above score is non-trivial as it involves integrating over parameters. Recently, \cite{yang23bayesian} showed that this score can be computed tractably and exactly for deterministic PCs. They employed the structure score together with the greedy cutset learning algorithm \cite{rahman2014cutset} as well as structural EM to learn state of the art PCs. They also showed that when the data is discrete, and the prior over the sum node parameters is assumed to be a Dirichlet distribution, the Bayesian Structure Score reduces to the well known Bayes-Dirichlet (BD) score \cite{heckerman1995learning}.

\subsection{Deep PCs via Random Structures}
Tangential to the above discussed direction of enhancing probabilistic circuits by learning their structure from data, there exists a diverse range of approaches that adopt the perspective that \textbf{structure may not be highly relevant when you have the ability to over-parameterize}. \cite{peharz20a-rat-spn} showed that by utilizing a sufficiently large ensemble of random structures, comparable performance can be achieved as with a learned structure. To create valid random structures, they extended the notion of Region Graphs introduced in \cite{ventura12buildspn,peharz2013greedy} to Random Region Graphs. Intuitively, a random region graph over a set of variables can be viewed as a rooted directed acyclic graph that recursively and randomly partitions the variables associated with each node. The region graph can be converted into a valid tensorized probabilistic circuit, called RAT-SPN \cite{peharz20a-rat-spn} by populating them with arrays of sum nodes, product nodes and leaf distributions.
 Unlike classical parameterizations, the  computational graph for the above parameterization has reduced sparsity, and is well amenable to GPU level-parallelization, making it highly scalable. Along similar lines, efficiently utilizing randomized structures for the specific case of deterministic probabilistic circuits was also explored by \cite{di2017fast,mauro21arandom}. Notably, \cite{peharz_20_einsum} extended the RAT-SPN framework by introducing a novel implementation design. They combined the vectorized sum and product operations into a single monolithic \textit{einsum} operation. This enabled designing PCs by stacking einsum layers similar to a deep neural network, allowing for even more parallelized computation, resulting in up to two orders of magnitude improvements in training times. Extending further, \cite{mari2023unifying} presented a unified framework that encompasses multiple such tensorized parameterizations for PCs, and showed that low-rank decompositions can be employed to improve their computational efficiency, while retaining expressivity.

\paragraph{Probabilistic Circuits vs Neural Networks.} As PCs are computational graphs, they are similar in essence to neural networks. The random parameterizations discussed above have enabled building deep PCs, bringing them closer to deep neural models. However, it is important to note that they still differ from neural networks on multiple aspects. Most importantly, the computational graph of a PC transforms probability densities associated with the data, unlike neural networks (and DGMs) which transforms the data itself. Further, the parameters associated with a PC have probabilistic semantics and as a result more structure in contrast to unconstrained parameters typical in the context of neural networks.

\subsection{Extensions and Modifications}
 Several attempts have been made to extend the definition of PCs for improving their expressivity and robustness.  \cite{sharir2018sum} proposed the introduction of quotient nodes representing conditional distributions within PCs. They showed that the resulting class of models are more expressive, while still capable of tractable inference. \cite{trapp2019bayesian} generalized sum nodes in a PC to mixtures with an infinite number of components. \cite{maua2017credal,maua2018robustifying} proposed replacing the scalar sum weights in a PC with intervals, resulting in robust circuits that represent a credal set of distributions. \cite{loconte2024subtractive} relaxed the non-negativity assumption on the sum node weights in a PC to build deep \emph{subtractive mixture models}. They showed that by squaring such models, we can construct valid PCs that are exponentially more expressive than their additive counterparts with convex sum node weights.
 
 Other approaches have tried to improve the expressivity by introducing more flexible leaf distributions. \cite{molina2018mixed} proposed the use of piecewise polynomial 
 leaf distributions. Similarly, \cite{trapp2020deepgaussianprocess} proposed the integration of gaussian processes at the leaves, however its added expressivity came at the cost of tractability. \cite{yu2024characteristic} proposed the use of univariate characteristic functions in the spectral domain as leaves in a PC to construct \emph{characteristic circuits}. They showed that the resulting model could better capture heterogenous data distributions that do not have closed form probability density functions.
 Figure \ref{pc-taxonomy} provides a systematic categorization of the various methodologies discussed above for improving PCs via better learning algorithms and design extensions.

\definecolor{header}{RGB}{250, 235, 220}
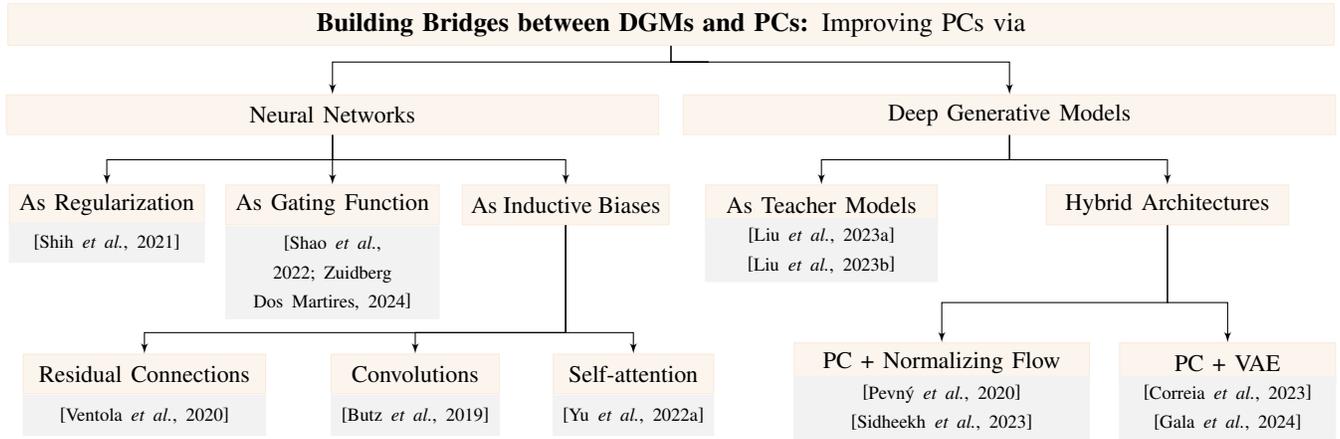
\begin{figure*}
    \centering
\begin{tikzpicture}[
    block/.style={
    rectangle,
    draw=header,
    fill=header!50,
    text centered,
    minimum height=1.5em},
    leaf/.style={
    rectangle,
    draw=gray!10,
    fill=gray!10,
    text centered,
    minimum height=1.5em},
    line/.style={draw, -latex'}
]
\node [block, fill=white!5, text width=0em,  minimum height=0em, draw=white] (anchor1)  at (0,0) {};
\node [block, text width=49.5em] (root)  at (7.5,0) {\textbf{Building Bridges between DGMs and PCs:} Improving PCs via};
\node [block, text width=24em] (nn) at (3,-1.2) {\small Neural Networks};
\node [block, text width=24em] (dgms) at (12,-1.2) {\small Deep Generative Models};

\node [block,text width=6.7em] (regularization) at (0,-2.4) {\small As Regularization};
\node [block,text width=7.4em] (gating)  at (3,-2.4)  {\small As Gating Function};
\node [block] (inductive) at (6.1,-2.4)  {\small As Inductive Biases};

\node [leaf,text width=6.7em] (cite-regularization) at (0,-2.9) {\scriptsize \cite{shih2021hyperspn}};
\node [leaf,text width=7.4em] (cite-gating)  at (3,-3.3)  {\scriptsize \cite{shao2022conditional,Zuidberg2024probabilisticneural}};

\node [block, text width=8.1em] (distillation) at (9.5,-2.4) {\small As Teacher Models};
\node [block, text width=8.5em] (hybrid) at (14.1,-2.4) {\small Hybrid Architectures};

\node [leaf, align=center, text width=8.1em] (cite-distillation) at (9.5,-3.) {
\scriptsize \cite{liu2023scaling} \\
\scriptsize \cite{liu2023understanding}};

\node [block, text width=10.5em] (pcnf)  at (11.1,-4.5)  {\small PC + Normalizing Flow};
\node [block, text width=7.5em] (pcvae)  at (14.9,-4.5)  {\small PC + VAE};

\node [leaf, align=center, text width=10.5em] (cite-pcnf) at (11.1,-5.1) {
\scriptsize \cite{sptn} \\
\scriptsize \cite{sidheekh2023probabilistic}};

\node [leaf, align=center, text width=7.5em] (cite-pcvae) at (14.9,-5.1) {
\scriptsize \cite{correia2023continuous} \\
\scriptsize \cite{gala2024probabilistic}};

\node [block, text width=8.5em] (resnets) at (0.5,-4.65) {\small Residual Connections};
\node [block, text width=5.7em] (conv) at (4.1,-4.65)  {\small Convolutions};
\node [block, text width=5.4em] (attention) at (7.,-4.65)  {\small Self-attention};

\node [leaf,text width=8.5em] (cite-resnets) at (0.5,-5.2) {\scriptsize \cite{ventola2020residual}};
\node [leaf,text width=5.7em] (cite-conv)   at (4.1,-5.2)   {\scriptsize \cite{butz2019deep}};
\node [leaf,text width=5.4em] (cite-attention) at (7.,-5.2)  {\scriptsize \cite{yu2022sumproductattention}};


\draw [line] (root)  |- (8, -0.5) -|  (dgms);
\draw [line] (root)  |- (8, -0.5) -|  (nn);

\draw [line] (nn) |- (3, -1.8) -|  (regularization);
\draw [line] (nn) |- (3, -1.8) -| (gating);
\draw [line] (nn) |- (3, -1.8) -| (inductive);


\draw [line] (dgms)  |- (12., -1.8) -| (distillation);
\draw [line] (dgms)  |- (12., -1.8) -|  (hybrid);


\draw[line] (inductive) |- (6.1, -4.1) -|   (resnets);
\draw[line] (inductive) |- (6.1, -4.1) -|   (conv);
\draw[line] (inductive) |- (6.1, -4.1) -|   (attention);

\draw[line] (hybrid) |- (14.1, -3.7) -|   (pcnf);
\draw[line] (hybrid) |- (14.1, -3.7) -|   (pcvae);


\end{tikzpicture}
\caption{A taxonomy of the literature on building hybrid models to bridge the gap between deep generative models and probabilistic circuits}
\label{dgm-pc-taxonomy}
\end{figure*}

\section{Building Bridges between DGMs and PCs}
With efficient and scalable deep parameterizations available for learning PCs, as outlined above, it is natural to assume that we can improve their expressive power by building larger overparameterized models. However, \cite{liu2023scaling} demonstrated that scaling the parameters of a PC does not result in a corresponding performance improvement. PCs are still {\bf far from achieving the expressivity of DGMs}. Thus, there is a growing interest in fusing concepts and inductive biases from deep generative models within PCs, to build hybrid models that can balance the expressive power of DGMs with the computational tractability, robustness, stability, 
and interpretability of PCs,  which we outline next.

\subsection{Probabilistic Circuits with Neural Networks}
One of the prevalent reasons that contributed to the popularity of deep learning was its ability to incorporate inductive biases such as translation invariance for image data effectively by employing convolutions. \cite{butz2019deep} demonstrated that the sum nodes in a PC are similar in essence to convolutions, while product nodes resemble pooling operations employed by deep neural models. They formalized the properties that such neural operations need to satisfy in order to result in a valid PC, thus building a class of hybrid and deep yet tractable convolutional PCs. Along similar lines, \cite{yu2022sumproductattention} proposed the integration of the \textit{self attention mechanism}, that has made transformer based models popular, within PCs, while \cite{ventola2020residual} proposed the use of residual links, thus developing a probabilistic analog to ResNets \cite{he2016deep}. \cite{shih2021hyperspn} proposed the use of neural networks for making PCs robust to overfitting. They partitioned the sum node weights of a PC into multiple \textit{sectors}, learned a lower dimensional embedding for each sector, and used small neural networks to map the embeddings to the parameters of the PC. 
This can be viewed as a soft weight-sharing mechanism where multiple parameters are generated by a single neural network. The resulting PC thus has reduced degrees of freedom and better generalization. \cite{shao2022conditional} considered conditional PCs for structured output prediction tasks, which can be viewed as modeling the conditional distribution $P_{\theta}(\mathbf{Y}|\mathbf{X})$ over a set of targets $\mathbf{Y}$ and features $\mathbf{X}$, and proposed integrating neural networks as gating functions. More specifically, they computed the mixture weights of the PC as a function of the input features $\mathbf{X}$ via a neural network. They demonstrated that the resulting model not only had added expressivity while retaining tractable inference capabilities over the target variables $\mathbf{Y}$, but could also be used to effectively impose structure over DGMs. More recently, \cite{Zuidberg2024probabilisticneural} extended conditional PCs capturing  factorizations of the joint distribution as given by a partial order $\mathcal{\psi}$ defined over the set of random variables $\mathbf{X}$ by generalizing the scalar weights associated with a sum node, say defined over $X \in \mathbf{X}$, to be a neural network transformation of the ancestors of $X$ in $\mathcal{\psi}$. The resulting model, called \emph{probabilistic neural circuits}, can be viewed as deep, hierarchical mixtures of Bayesian Networks.

\subsection{Probabilistic Circuits with VAEs}

 \cite{liu2023scaling} attributed the failure of PCs in the overparameterized regime to the increase in latent information (associated with sum nodes) as PCs are scaled, which in turn made the marginal likelihood over observed variables more complex and hence maximum likelihood training more challenging. To address this issue, they proposed to provide extra supervision to PC learning by explicitly materializing the latent variables using a less tractable but more expressive deep generative model. As VAEs are effective models for learning latent representations, \cite{liu2023scaling} utilized Masked Autoencoders \cite{he2022masked} to learn feature representations for the sum nodes and employed K-means clustering in this feature space to obtain the assignments for the discrete latent variables associated with sum nodes. The resulting framework, which they called Latent Variable Distillation (LVD) was able to achieve competitive performance against widely used DGMs. Building further, \cite{liu2023understanding} studied the theoretical properties as well as design principles of the DGMs to be used as teacher models for LVD. They observed that when performing LVD, the performance of the student PC can exceed that of the teacher DGM. They also highlighted that the disparity between the continuous latent representations learned by DGMs and the discrete latent variable assignments demanded by a PC can lead to information loss. They proposed to overcome this issue by employing a progressive growing algorithm that leveraged feedback from the PC to perform dynamic clustering. 

VAEs can be interpreted as mixture models with an infinite number of components, where the components depend continuously on the latent code, and are thus intractable. PCs on the other hand are discrete hierarchical mixture models. 
\cite{correia2023continuous} observed that even a huge discrete mixture model such as an overparameterized PC is unable to outperform a relatively moderately sized uncountable mixture model such as a small VAE, suggesting that continuous mixtures generalize better or are easier to learn as opposed to PCs. Thus, they proposed to merge VAEs with PCs by taking continuous mixtures of tractable PCs. Their approach can be intuitively understood as replacing the decoder of a VAE with a PC. Though the resulting formulation is intractable in practice, the authors demonstrated that it can be approximated arbitrarily well using numerical integration techniques when the considered latent space has a low dimension. \cite{gala2024probabilistic} further generalized this approach by allowing integral units representing continuous latent variables to be defined not only at the root but also as internal nodes in a PC. 

In the reverse direction, improving VAEs by using PCs was explored by \cite{tan2019hierarchical}, who built hierarchical mixtures of VAEs by employing them at the leaf nodes of a PC. Though the resulting model lacked tractability, the authors showed that such a hybrid model resulted in improved stability and better learning  of VAEs. 

\subsection{Probabilistic Circuits with Normalizing Flows}
Normalizing flows constitute one of the most structured class of DGMs that utilize diffeomorphic neural transformations to map a simple base distribution into a more complex one. Their diffeomorphic structure enables computing the probability density exactly using the change of variables formula, and are thus tractable models for evidential inference. Naturally, extending PCs by employing the change of variables principle of flows has been explored in the literature. To achieve this, \cite{sptn} proposed the addition of a new  node type - called transform nodes, arbitrarily over existing nodes in a PC. Each transform node ($\mathcal{T}$) was associated with an invertible affine transformation and defined over a single child node. The output of $\mathcal{T}$ was defined recursively as the transformation of the distribution modeled by its child.

Building further, \cite{sidheekh2023probabilistic} showed that the above construction for integrating normalizing flows with PCs can violate the decomposability of PCs, hence making complex inference queries such as marginals and conditionals intractable. They formalized the necessary conditions that transform nodes need to satisfy in order to retain tractability, which they called $\tau-$decomposability. Intuitively, $\tau-$decomposability demanded that when $\mathcal{T}$ is defined over a product node $\mathcal{P}$, it should transform the scopes of the children of $\mathcal{P}$ independently. They also showed that integrating $\tau-$decomposable transform nodes arbitrarily in a circuit is equivalent to defining normalizing flows over the leaf distributions of a PC. They demonstrated that utilizing invertible linear rational spline transformations at the leaves, we can build expressive yet tractable \emph{probabilistic flow circuits}. 

\subsection{Implications of Expressive \& Tractable Models}

\paragraph{From Approximations to Exactness.}
The tractability offered by PCs can be utilized to solve classical problems solved via approximations \emph{exactly}. \cite{shih2020probabilistic} studied PCs in the context of discrete graphical models, and showed that they can be employed as expressive variational families that support exact ELBO computation as well as stable gradients. \cite{khosravi2019tractable} showed that PCs can be used to compute expectations as well as higher order moments of the predictions of discriminative models. 

\paragraph{To More Complex Tasks.}
\cite{choi2022solving} showed that the complex inference routine of Marginal MAP can be solved exactly by leveraging PC transformations. \cite{ventola2023probabilistic} showed that Monte Carlo Dropout, when introduced within the context of PCs and be used for uncertainty quantification exactly and efficiently, thus making PCs robust to out of distribution data.
\cite{selvam2023certifying} utilized the tractability of PCs to reason with partial data to search for discrimination patterns and certify the model's \textit{fairness}. 

\cite{vergari2021acompositional} compiled a comprehensive catalogue of operations invloving probability distributions that can be tractably computed using PCs, outlining the structural properties that needs to be satisfied within the context of each operation. Their work generalized the common inference queries we have seen so far to also include the computation of sums, products, quotients, powers, logarithms and exponentials of probability distributions encoded as PCs. As a result, complex information theoretic quantities such as divergence measures, which typically demand approximations, could be represented via tractable and modular operations over circuits.

\paragraph{Broader Applications Across Fields.} PCs, as expressive and tractable models, have found applications across various domains spanning causality, neuro-symbolic learning, and controlled generation. Notably, \cite{wang2023compositional} showed that compositions of probabilistic operators defined via PCs can be used to answer causal inference queries such as the \emph{backdoor adjustment} in polynomial time. \cite{liu2022lossless} showed that PCs can be adapted to perform lossless compression effectively. \cite{ahmed2022semantic} showed that PCs can be used to restrict the support of deep classifier outputs, serving as an effective interface to integrate symbolic knowledge with deep neural models. Further, PCs have also been employed to effectively control the generative process in DGMs such as autoregressive language models \cite{zhang2023tractable} and diffusion models \cite{liu2024image}. 

\section{Open Problems and Promising Directions}
We have so far covered the various algorithms and design principles that has enabled building expressive and tractable generative models.
However, the field has several open problems, which makes it a fertile area for further research and significant advancements in several directions. 

\paragraph{Theory of Optimizing Overparameterized PCs.} Most works that attempt to overcome the performance plateau of overparameterized PCs are based on heuristics. A theoretically grounded understanding of this phenomenon is yet to be developed. On the other hand, overparameterization in the context of neural networks is well studied. For e.g., the phenomenon of double-descent has been well studied for overparameterized neural networks, but not explored in PCs. Thus, borrowing such notions to understand the characteristics of PC loss landscapes and building more efficient optimizers that can make use of the tractability of PCs presents a promising avenue for future research.

\paragraph{Latent Representation Learning.} Learning semantically meaningful and disentangled latent representations is a fundamental goal in generative modeling. The sum nodes in a PC introduces probabilistically meaningful latent variables. However, learning useful data respresentations using them is non trivial and less explored. Indeed, the works \cite{vergari2018sum,vergari2019visualizing} have laid some of the foundations in this direction by extracting interpretable representations from a PC by looking at node activations. However a scalable and differentiable approach to learning such representations is still lacking. The recent works \cite{liu2023scaling,liu2023understanding} have enabled distilling information from VAEs within PCs. Extending this framework to support representation learning for PCs is a promising future direction. 

\paragraph{Adversarial Training.} Maximum likelihood learning employed for training PCs, though stable, is known to achieve suboptimal sample quality. Prior works in the context of DGMs have thus explored augmenting the MLE objective with adversarial losses to improve sample generation.  While recent works \cite{peddi2022robust} have studied the robustness of PCs to adversarial attacks, utilizing adversarial losses is relatively less explored. A key challenge here is that sampling in PCs is typically non-differentiable. Hence, unlike GANs, backpropagating the output of an adversarial discriminator for the generated samples is difficult. However, recent works \cite{shao2022conditional,lang2022elevating} have explored differentiable sampling strategies for PCs, making adversarial training of PCs a potential way to improve their expressivity. 

\paragraph{Incorporating Symmetries.} Enabling the adoption of  hybrid probabilistic models in real world applications demand that we instill in them the ability to model domain specific inductive biases. While such designs have been explored for domains such as images \cite{butz2019deep}, time series \cite{yu2022predictive}, and tree structured graphs \cite{papez2024sumproductset}, extending them to capture the symmetries, invariances, and equivariances needed for domains involving relational data, sets, and general graphs is an area of active research.

\paragraph{Multi-Modal Learning.} With the increasing abundance of heterogenous data, building PCs that can utilize multiple (if not all) modalities of available data to make effective and reliable decisions is an important and open research problem. Recent works on integrating flows with circuits \cite{sidheekh2023probabilistic} have laid the foundations for modeling flexible leaf distributions within PCs. One way to enable probabilistic multi-modal and perhaps compositional learning would thus be to embed normalizing flows trained over different modalities as leaf distributions in PCs. 

\paragraph{Applications in Other Fields.} Finally, expressive probabilistic models can be used as replacements for DGMs in various learning paradigms. For instance, one of the applications of DGMs within reinforcement learning involves their use as world models \cite{ha2018recurrent}, which enables an agent to hallucinate the behavior of its environment to take better actions. Employing PCs in such a context could additionally empower the agent to perform probabilistic inference over the environment dynamics. Another example is \emph{active feature acquisition}, where DGMs have been employed to assess potential information gain associated with acquiring new features \cite{li2021active}. A PC, when employed in this context, brings with it the power to exactly and efficiently compute information theoretic quantities of interest and reason over relevant subsets of features via marginalization. 

\section{Conclusion}
We presented an extensive overview of the current research on tractable probabilistic models, focusing on PCs and discussing the various algorithmic and design extensions aimed at improving their expressivity. We also outlined how recent works have attempted to bridge the gap between the expressivity of DGMs and the tractability of PCs by building hybrid models. 
An important point to highlight is that 
\textit{structure} is  an inevitable factor when learning probabilistic generative models. Stabilizing the training of DGMs such as GANs and VAEs often requires imposing a weak structure over their parameters. The invertible structure imposed by normalizing flows enables exact density evaluation and stable maximum likelihood training.  Imposing more restrictive classes of structure over PCs helps us gain tractability over increasingly complex queries. The key focus thus should be to build generative models that can exploit the right level of structure required for solving the tasks at hand. By understanding the design principles of generative modeling and merging them to build hybrid models, we gain the ability to interpolate on this tractability-expressivity spectrum. The possibilities that await such flexible probabilistic models are virtually infinite.

\section*{Acknowledgements}
The authors gratefully acknowledge the generous support by the AFOSR award FA9550-23-1-0239, the ARO award W911NF2010224 and the DARPA Assured Neuro Symbolic Learning and Reasoning (ANSR)
award HR001122S0039. 

\bibliographystyle{named}
\bibliography{ijcai24}

\end{document}